\documentclass[]{spie}  

\usepackage{amsmath,amsfonts,amssymb}
\usepackage{graphicx}
\usepackage[colorlinks=true, allcolors=blue]{hyperref}
\usepackage{float}
\usepackage{comment}

\title{Radiology Report Generation Using Transformers Conditioned with Non-imaging Data}

\author[a,c]{Nurbanu Aksoy}
\author[a,c*]{Nishant Ravikumar}
\author[a,b,c,*]{Alejandro F Frangi}
\affil[a]{CISTIB Centre for Computational Imaging and Simulation Technologies in Biomedicine}
\affil[b]{Leeds Institute for Cardiovascular and Metabolic Medicine, School of Medicine}
\affil[c]{School of Computing, University of Leeds, Leeds, UK}
\affil[*]{Indicates joint last authors}

 \authorinfo{Further author information: (Send correspondence to N.A)\\
 N.A.: E-mail: N.Aksoy1@leeds.ac.uk \\
 N.R.: E-mail: N.Ravikumar@leeds.ac.uk \\
 A.F.: E-mail: A.Frangi@leeds.ac.uk, Telephone: +44(0)113 343 9640}

\begin{document} 
\maketitle

\begin{abstract}
Medical image interpretation is central to most clinical applications such as disease diagnosis, treatment planning, and prognostication. In clinical practice, radiologists examine medical images (e.g. chest x-rays, computed tomography images, etc.) and manually compile their findings into reports, which can be a time-consuming process.
Automated approaches to radiology report generation, therefore, can reduce radiologist workload and improve efficiency in the clinical pathway. While recent deep-learning approaches for automated report generation from medical images have seen some success, most studies have relied on image-derived features alone, ignoring non-imaging patient data. Although a few studies have included the word-level contexts along with the image, the use of patient demographics is still unexplored. On the other hand, prior approaches to this task commonly use encoder-decoder frameworks that consist of a convolution vision model followed by a recurrent language model. Although recurrent-based text generators have achieved noteworthy results, they had the drawback of having a limited reference window and identifying only one part of the image while generating the next word. 
This paper proposes a novel multi-modal transformer network that integrates chest x-ray (CXR) images and associated patient demographic information, to synthesise patient-specific radiology reports. The proposed network uses a convolutional neural network (CNN) to extract visual features from CXRs and a transformer-based encoder-decoder network that combines the visual features with semantic text embeddings of patient demographic information, to synthesise full-text radiology reports. The designed network not only alleviates the limitations of the recurrent models but also improves the encoding and generative processes by including more context in the network. 
Data from two public databases were used to train and evaluate the proposed approach. CXRs and reports were extracted from the MIMIC-CXR database and combined with corresponding patients' data (gender, age, and ethnicity) from MIMIC-IV. Based on the evaluation metrics used (BLEU 1-4 and BERTScore), including patient demographic information was found to improve the quality of reports generated using the proposed approach, relative to a baseline network trained using CXRs alone. The proposed approach shows potential for enhancing radiology report generation by leveraging rich patient metadata and combining semantic text embeddings derived thereof, with medical image-derived visual features. 

\end{abstract}


\section{INTRODUCTION}
\label{sec:intro}  
In almost all branches of health sciences, medical imaging modalities are used for disease diagnosis, accurate treatment planning, patient care and prognosis. Interpretation of medical images is undertaken by radiologists, and their findings are compiled into a full-text report, taking into account other relevant patient information (such as clinical history, and patient demographics). It is important that these reports are comprehensive, accurate and generated in a short period of time, with a specific pre-defined structure.
In clinical practice, chest X-ray (CXR) images are the most widely used imaging modality and it is generally the first step in screening patients for a large variety of lung diseases. Reports generated following CXR examinations typically include radiologists' interpretations categorised as 'findings', and 'impressions', and are indicative of normal and abnormal features/appearances in image regions. Writing such extensive reports requires comprehensive domain knowledge and experience. Additionally, it is a time-consuming, laborious, and error-prone task, even for experienced staff. Given the large volume of CXR examinations performed regularly in most hospitals, approaches that facilitate the automatic generation of radiology reports could help reduce radiologists' workload and improve efficiency in clinical pathways.

In this context, there have been several attempts to automatically generate radiology reports. Most existing deep learning approaches proposed for radiology report generation for given CXR images, leverage networks comprising a convolutional encoder and recurrent decoder, which was originally introduced for the task of image captioning\cite{monshi2020deep,boag2020baselines,kaur2021methods}  The majority of existing literature on report generation that is based on deep learning methods, leverage the convolutional encoder-recurrent decoder type frameworks that were originally introduced in image captioning task \cite{monshi2020deep,boag2020baselines,kaur2021methods}. Recently, due to the great success of transformers in neural machine translation tasks, a few studies have been shifting from recurrent models to transformer-based models \cite{xiong2019reinforced,cornia2020meshed,chen2020generating,nooralahzadeh2021progressive}.
Although previous studies have shown promising results, they often treat the task of report generation as an image captioning problem. While radiology report generation shares some similarities with image captioning, it differs in other respects, such as: (1) the annotations for medical images (e.g. CXR images) are in the form of paragraphs rather than a single (short) sentence and correspondingly, the generated reports are required to be more comprehensive than expected in typical image captioning tasks encountered in the computer vision domain; (2) distinctions between medical images are subtle and visual semantic information is not as easy to extract as for natural images; and (3) the information needed to generate image captions is often embedded in natural images, however, additional information/context may be required to accurately analyse and interpret medical images.

Existing approaches for CXR report generation utilise just the images as inputs and ignore additional non-imaging patient information that is typically available to radiologists when interpreting images. Learning from CXR images alone is challenging as each report is originally generated based on relevant medical records and information (such as gender, age, weight, pre-existing conditions/comorbidity information, etc.). As CXR images are two-dimensional projections of three-dimensional structures, some useful information is lost and semantic gaps occur in the data available to networks/algorithms to learn from. Some studies have tried to break the resulting semantic gap by adding more context to the network. They have extracted some medical concepts from the target sequence \cite{yuan2019automatic}, produced a high-level context for each report \cite{nooralahzadeh2021progressive}, classified the images into different categories \cite{alfarghaly2021automated}, or classified the reports into normal and abnormal classes \cite{singh2021show}. While these approaches have demonstrated some success, they primarily rely on augmenting information seen by the network and do not enrich the context provided to the network during training with new information (such as patient demographic data for example).

With the aim of addressing the existing semantic gap, we introduce a novel multi-modal transformer-based deep neural network that is capable of extracting information from imaging and non-imaging patient data simultaneously, to generate radiology reports. To the best of our knowledge, this is the first study to explore automatic CXR report generation by combining information from CXR images with non-imaging structured patient data (i.e. not available for linked radiology reports).

\section{RELATED WORKS}

\subsection{Visual Captioning Methodologies}

Visual Captioning, also known as Image Captioning, is the task of automatically generating a natural language description of a given image. Firstly, it requires the recognition and detection of objects located in the image, the identification of attributes, and the determination of their relationships and interactions. Subsequently, it must generate coherent sentences based on the features extracted. Since the task entails the incorporation of computer vision and natural language processing, it has attracted tremendous attention in the artificial intelligence community. 
Based on the recent literature, the most common deep learning architecture used by researchers follows a standard encoder-decoder baseline that consists of two phases. Broadly, the encoder part receives an image and sends it to Convolutional Neural Network (CNN) which is usually pre-trained on large datasets for classification and recognition tasks. The layers of CNN extract region-based visual features and these high-level image representations are used as input by Recurrent Neural Network (RNN)-based decoder to generate a relevant caption.

However, the recurrent-based report generation models have a well-known vanishing and exploding gradients problem which means that the recent input sequence causes a bias as there is limited access to the previous inputs and there  is no direct access to all inputs. Leveraging recent advances in transformers \cite{vaswani2017attention}, which is a self-attention-based neural network, in the natural language processing (NLP) area, state-of-the-art image captioning architectures have tended to substitute their model components with the transformers \cite{liu2021cptr}. The main advantages of the transformers over other architectures are that it does not use recurrence, and it is entirely based on an attention mechanism. It takes and executes the input sequence as a whole, allows more parallelisation, and learns the relationship between words in the sequences by the use of multi-head attention mechanisms and positional embeddings. Since more context is included in the network, transformer-based architectures can learn faster and more effectively.
In the wake of this phenomenon, several variants of transformers were developed for image caption generators. The original transformer architecture also inherits an image-text embedding type framework which consists of an encoder and decoder. 

The most commonly adapted transformer-based encoder-decoder architecture for image captioning has three main components; visual feature extraction model, Transformer-based encoder and transformer-based decoder. As in previous studies, pre-trained CNN models are also employed for high-level feature extraction. However, in this approach, the output of the visual model is used by a transformer-based encoder to map the visual features and generate the sequence of image representations. Then, the transformer-based decoder receives the results of the encoder to generate a corresponding caption of the given image.
\cite{zhu2018captioning} presented Captioning Transformer (CT) model that uses ResNeXt\cite{xie2017aggregated} CNN model as an encoder and Transformer as a decoder. \cite{zhang2019image} also used the Transformer model as a decoder along with the ResNet CNN model, additionally, they improved the network with a combination of spatial and adaptive attention. \cite{li2019entangled} enhanced the vanilla Transformer architecture with Entangled Attention (ETA) and Gated Bilateral Controller (GBC), their proposed model allows the process of semantic and visual concepts concurrently.

Moreover, \cite{cornia2020meshed} introduced a fully-attentive model called a Meshed-Memory Transformer that consists of a Memory-Augmented Encoder, which has enriched with learnable the keys and values with a priori information and used a learnable gating mechanism to perform mesh connectivity and Meshed Decoder that performs a meshed connection between all encoding layers. On the other hand, \cite{liu2021cptr} proposed a full Transformers network without having any convolutional operation in the encoder. Different from previous studies, their model, CaPtion TransformeR (CPTR), uses the raw image and adjusts it according to the accepted input form of the transformer encoder by dividing the original image into N patches. After reshaping the patches, the obtained patch embedding is incorporated with positional embedding.

\subsection{Radiology Report Generation}

In recent years, many studies have had great success in fine-tuning deep neural networks to generate medical reports. Most existing radiology report generation studies adopt image captioning approaches for medical report generation and leverage the CNN-LSTM framework; \cite{jing2017automatic} employed a pre-trained VGG-19 model to learn visual features and use the extracted features to predict relevant tags for any given chest X-ray. The predicted tags are used as semantic features in the network and both semantic and visual features are fed into the Co- Attention Network. Hierarchical LSTM has used the context vector provided by Co-attention Network to generate the topic and description of the given X-ray. Although the model obtained promising results and achieved great success in its field, the repetitive sentences in reports and the generation of different results for the same patient undermined its credibility from both medical and computational perspectives. \cite{xue2018multimodal} improved the pre-trained Resnet- 152 encoder using multi-view content (both lateral and frontal view) and incorporated them to ensure the consistency of the results. They also generated a report with a sentence decoder and additionally used the first predicted sentence as a joint input along with image encoding.

Another major study was proposed by \cite{yuan2019automatic}, who pre-trained their multi-view encoder from scratch using the CheXpert dataset \cite{irvin2019chexpert} instead of using ImageNet pre-trained models. In order to enhance the decoder, they extracted and applied medical concepts from the reports. Applying medical concepts conveyed the semantics in the content of the report and they achieved noteworthy results. The idea of using medical concepts was also employed by \cite{yang2020automatic}; whilst they applied a similar approach in principle, they proposed a reward term to extract more precise concepts. Although the medical concepts obtained were more accurate compared with other studies, they were still not very informative about the given X-Ray.

On the other hand, \cite{singh2021show} argued that the format of the normal and abnormal reports differs, therefore, a single framework cannot handle both styles accurately. To overcome this limitation, they followed a slightly different approach and first, they classified the reports as normal and abnormal. Then, they generated the Findings section and summarised it to acquire the Impression section for both normal and abnormal reports. They also adopted pre-trained CNN model, InceptionV3, for visual feature extraction and used attention-based LSTM for text generation.
More recently, studies have taken advantage of Transformer for medical report generation, after its success for text generation based on non-linguistic representation. \cite{xiong2019reinforced} designed a hierarchical Transformer model which contains a novel encoder that can extract the regions of interest from the original image by using a region detector and uses these regions to obtain visual representations. Moreover, \cite{chen2020generating} introduced a medical report generator via a memory-driven Transformer. They have used a relational memory to keep the knowledge from the previous case, in this manner, the generator model can remember similar reports when generating the current report. \cite{nooralahzadeh2021progressive}  proposed a progressive Transformer-based report generation framework that produces high-level context from the given X-ray and converts them into a radiology report by employing the Transformer architecture. Their proposed model consists of pre-trained CNN as a visual backbone, a mesh-memory Transformer \cite{cornia2020meshed} as a visual language model and BART \cite{lewis2019bart}  as a language model.
\section{DATA AND METHODS}

\subsection{Data and Pre-processing}
\label{sec:data_process}

Two publicly available databases, namely, MIMIC-CXR \cite{johnson2019mimic} and MIMIC-IV \cite{mimiced} were used to create the dataset used throughout this study for training and evaluating the proposed approach.

\textbf{MIMIC-CXR (version 2.0.0)} consists of 377,110 CXR images (anteroposterior, posteroanterior, and lateral views) and the corresponding 227,835 de-identified free-text radiology reports, acquired from 63,473 patients. Each report has several sections including - 'examination', 'indication', 'technique', 'comparison', 'findings', and 'impressions'. However, there is a lot of missing information in both the 'comparison' and 'indication' sections due to the anonymisation of the data. Moreover, the 'impressions' section has several identical entries across different patients, resulting in data imbalance/biases. Therefore, only the 'findings' section of the reports is used throughout this study. All reports were subsequently pre-processed and standardised in the following way - the length of each report was calculated and reports containing less than 9 words were removed from the dataset. All characters in all reports were converted into lowercase. Punctuation marks, tokens with a number, stop words, and sequences that do not provide meaningful information were removed from each report. Also, as we do not have access to patients' historical clinical records, we removed the reports referring to prior studies on the same subject. Then, different texts that convey the same meaning (i.e. are semantically similar) were standardised to minimise variability. This step is neglected by the other studies. However, it is important to reduce the diversity of the language and terminology used in the reports. 
Finally, $<$start$>$ and $<$end$>$ tokens were added to each sequence. On the other hand, the images were filtered to only include anteroposterior and posteroanterior projections. Images were then re-sized to 299 px × 299 px and their intensities were normalised.

\textbf{MIMIC-IV (version 1.0)} consists of anonymised patient information who were admitted to the Beth Israel Deaconess Medical Center (BIDMC). Firstly, gender, age, and ethnicity values for each patient were extracted from relevant tables. The gender values were converted into binary data by assigning a value of 0 for female and 1 for male patients. We only used the five ethnicity values that occurred the most, and we created a new variable for each of them using one-hot encoding. The range of age was restricted to 19 to 91 and then normalised to be between 0 and 1. Both MIMIC-CXR and MIMIC-IV databases have a common unique identification number for each patient, which is used as a key to retrieve relevant information and create a data point for each subject. The final dataset has 43,628 data points, and each data point has a CXR image, the corresponding findings section/report, and gender, age, and ethnicity data. One of the main challenges for the report generation task is the biased dataset. In addition to the fact that most of the cases are normal, the several instances of identical reports across different patients pose additional challenges. To address this problem, we sampled the dataset and created 4 subsets that were as balanced as possible, which were used for the experiments conducted in this study. We split each subset into training, validation and test sets with a split ratio of 70: 20: 10, respectively.

\subsection{Network Design}
\begin{figure} [H]
   \begin{center}
   \begin{tabular}{c} 
   \includegraphics[width=0.7\textwidth]{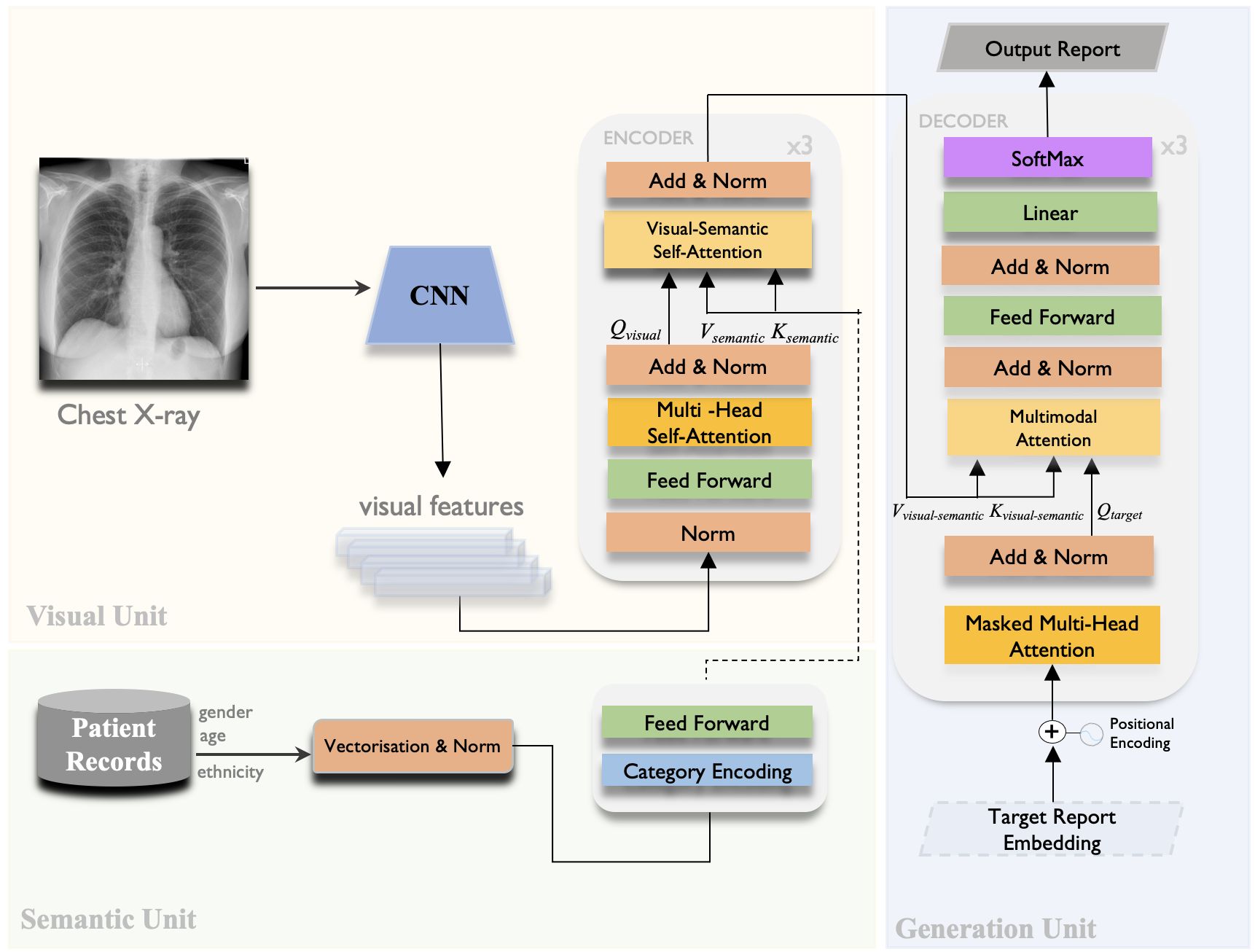}
   \end{tabular}
   \end{center}
   \caption[model] 
   { \label{fig:model} 
The overall framework of the proposed CXR report generation model.}
   \end{figure}

As shown in Fig.~\ref{fig:model}, the report generation network has three main components: Visual Unit, Semantic Unit, and Generation Unit. The representation of each image is derived in two steps in the \textbf{Visual Unit}. The \textbf{Visual} and \textbf{Semantic Units} constitute the encoder and the \textbf{Generation Unit} represents the decoder in the proposed approach. First, the image is passed through a CNN (EfficientNet was used) for visual feature extraction. The pre-trained model receives resized (299x299) images and generates a 1280-length visual feature vector. The extracted features are then used in the Multi-Head Self-Attention block for deeper understanding of the relationships between the detected entities. In the \textbf{Semantic Unit}, meanwhile, the relevant patient data (gender, age and ethnicity) are encoded  and semantic features are obtained. In order to enhance the image representation, we add a Visual-Semantic Self-Attention block to the encoder which uses both types of features and derives a new hybrid image representation. This hybrid image representation is in turn provided as input to the decoder. In the decoder/\textbf{Generation Unit}, the target report corresponding to the given CXR image is converted into a vector representation with positional encoding and fed to the transformer-decoder (during training only), which in turn generates the target report, conditioned on the derived hybrid image representation. 

\subsection{Implementation Details}

In the encoding phase, we employ EfficientNet \cite{tan2019efficientnet} as a base model to extract visual features and it is trained end-to-end by fine-tuning all parameters. The pre-trained model receives resized (299x299) images and generates a 1280-length visual feature vector. Following normalisation, the image vector is passed through the feed-forward layer with the ReLU activation function. The output of the feed-forward layer is sent to the multi-head self-attention (MHA) layer followed by the normalisation layer. The normalised image representation is used as a query parameter in the Visual-Semantic Self-Attention block. Meanwhile, the relevant patient data (gender, age and ethnicity) are extracted from the MIMIC-IV database, After the semantic data is vectorised and encoded as described in Sec.~\ref{sec:data_process}, it is passed to a fully connected layer. During the experiments, we first presented each patient data type to the model alone, and then we performed further experiments to concatenate the types that showed improvements.

The decoder is an auto-regressive model that generates words recursively, starting with the $<$start$>$ token associated with each tokenised target report and stops decoding/predicting words when the $<$end$>$ token is generated. Due to its auto-regressive nature, the decoder takes the information from the previous iteration to predict the next word. During training, the target report corresponding to the given image is passed through an embedding layer and used as input to train the decoder. The first layer of the decoder is a masked multi-head self-attention layer. Different from MHA in the encoder, the masking method is used here to ensure that the attention mechanism does not share any information about future tokens. In this way, each token only has access to information regarding itself and the previously generated values. The output of the masked multi-head attention layer is combined with the output of the decoder using a multi-modal attention layer, where the query matrix is created from the former and semantically enhanced hybrid image representations are used for the keys and values matrix. The output of the decoder is passed through a SoftMax layer, which calculates probability scores between 0 and 1, and returns the word with the highest probability score as the predicted word. During inference, only CXR images and relevant patient data are used as inputs and the report generation process is triggered with $<$start$>$ token. The decoder generates words in a recursive loop until the $<$end$>$ token is predicted. 

We built five different models; the network trained on CXRs alone is used as a baseline model. We named the other models based on the type of patient data used in the network. They adopt the same principles to combine the non-imaging data with image-derived features. Each model was trained using the Adam optimiser (with a learning rate of 3e-4) and a Sparse Categorical Cross Entropy loss function. A batch size of 64, a maximum report length of 50, a vocabulary size of 2212, a dense dimension of 512, and an embedding dimension of 512 are used during training and validation.

The decoder begins with $<$start$>$ token, generates words recursively and stops the decoding process when the $<$end$>$ token is generated. Due to its auto-regressive nature, it takes the information from the previous iteration to predict the next word. During training, the target report corresponding to the given image is passed through the embedding layer. The transformer does not implement a loop and all inputs are processed in parallel. Although this is one of its main advantages over recurrence models, the notion of the sequence order is lost during this operation. Therefore, positional information is injected into the output of the embedding layer with the positional encoding and it is sent to the first layer of the decoder which is a masked multi-head self-attention layer. Different from MHA in the encoder, the masking method is used here to ensure that the attention mechanism does not share any information about future tokens. In this way, each token only has access to information regarding itself and the previously generated values. The MHA layer is followed by the normalisation layer and Multi-modal attention layer, respectively. The multi-modal attention layer incorporates information from both the encoder and the decoder. When calculating self-attention, the queries matrix is created from the previous layer, and the semantically enhanced image representations are used for the keys and values matrix. The output goes through the normalisation layer followed by the feed-forward layer with the ReLU activation function and the last feed-forward layer takes the outputs and functions as a classifier. The final output is passed through the SoftMax layer, which calculates probability scores between 0 and 1, and returns the word with the highest probability score as the predicted word. The model was trained using the Adam optimiser (with a learning rate of 3e-4) and a Sparse Categorical Cross Entropy loss function. A batch size of 64, a maximum report length of 50, a report vocabulary size of 2212, a dense dimension of 512, and an embedding dimension of 512 are used during training and validation. During inference, only CXR and relevant patient data are used and the report generation process is triggered with $<$start$>$ token. The decoder generates the words in a recursive loop until the $<$end$>$ token is predicted. We also define a text sampling function that selects candidate words from a probability array and set them to a temperature parameter is 0.5.

\section{RESULTS and DISCUSSION}
Following data pre-processing (described in Sec.~\ref{sec:data_process}), the total data size was reduced to 40,700. We, then, sampled four different sets of 4500 data with no overlap. Each model investigated was trained and evaluated on these sets using the same train, validation and test split for a fair comparison. The average results of the models across all four test sets are summarised in Table~\ref{tab:results} using the Bilingual Evaluation Understudy (BLEU) score and BERTScore. 
Diversity of the language has an effect on the BLEU score since it measures the similarity between target text and ground truth based on n-gram precision. As it is insufficient to capture semantic similarity, we also evaluated the models with BERT-Score that leverages pre-trained BERT contextual embeddings in order to calculate the token similarity instead of relying on exact string matching.

The results show that the use of patient data in the report generation task is beneficial and improves the model performance. We also assessed the statistical significance of the obtained results using the Student's t-test with a significance level of $\alpha$ = 0.05 for each metric. The best results showing significant improvements over the rest are indicated in bold in Table~\ref{tab:results}. The ethnicity-enriched model achieved the highest scores followed by the gender- and age-enriched models.

\begin {table}[ht!]
\caption{Quantitative results using Natural Language Generation metrics } 
\label{tab:results}
\centering
\resizebox{\textwidth}{!}{\begin{tabular}{|l|l|l|l|l|l|l|l|} 
\hline
\rule[-1ex]{0pt}{3.5ex} Model & BLEU-1 & BLEU-2 & BLEU-3 & BLEU-4 & $
$ & $R_{BERT}$ & $F1Score_{BERT}$ \\

\hline 
\rule[-1ex]{0pt}{3.5ex} Baseline Model & 0.314 $\pm$ 0.004  & 0.193 $\pm$ 0.003 & 0.140 $\pm$0.001 &0.089 $\pm$ 0.002 &0.27 $\pm$ 0.003 &	0.20 $\pm$ 0.002 &0.23 $\pm$ 0.03\\ 

\hline
\rule[-1ex]{0pt}{3.5ex} Gender Model   & 0.322 $\pm$ 0.003 &  0.209  $\pm$ 0.004&\bfseries0.142 $\pm$ 0.002  &	\bfseries 0.091 $\pm$ 0.002 & 0.27 $\pm$ 0.001 &\bfseries0.24 $\pm$ 0.003 &	0.25 $\pm$ 0.002  \\
\hline
\rule[-1ex]{0pt}{3.5ex}  Age Model     &0.322 $\pm$ 0.002 &   0.205 $\pm$  0.005  &	   0.141  $\pm$ 0.003 &	0.088  $\pm$ 0.003 &0.28  $\pm$ 0.001 &	0.21 $\pm$ 0.002 &	0.25  $\pm$ 0.004 \\

\hline
\rule[-1ex]{0pt}{3.5ex} Ethnicity Model & \bfseries 0.333 $\pm$ 0.004	& \bfseries0.210 $\pm$ 0.003 &  \bfseries 0.142 $\pm$ 0.002 &  0.088 $\pm$ 0.002 & \bfseries	0.30 $\pm$ 0.013 & 0.23 $\pm$ 0.008 & \bfseries 0.27 $\pm$ 0.010\\
\hline
\rule[-1ex]{0pt}{3.5ex} Ethnicity+Gender & 0.321 $\pm$ 0.003 &	 0.206 $\pm$ 0.005&  0.141 $\pm$ 0.002 & 0.087 $\pm$ 0.002  & 0.26 $\pm$ 0.009  & 0.22 $\pm$ 0.005 &	0.24 $\pm$ 0.006  \\
\hline
\end{tabular}}
\end{table}

\newpage

\begin{figure} [H]
\begin{center}
\begin{tabular}{c} 
\includegraphics[width=0.7\textwidth]{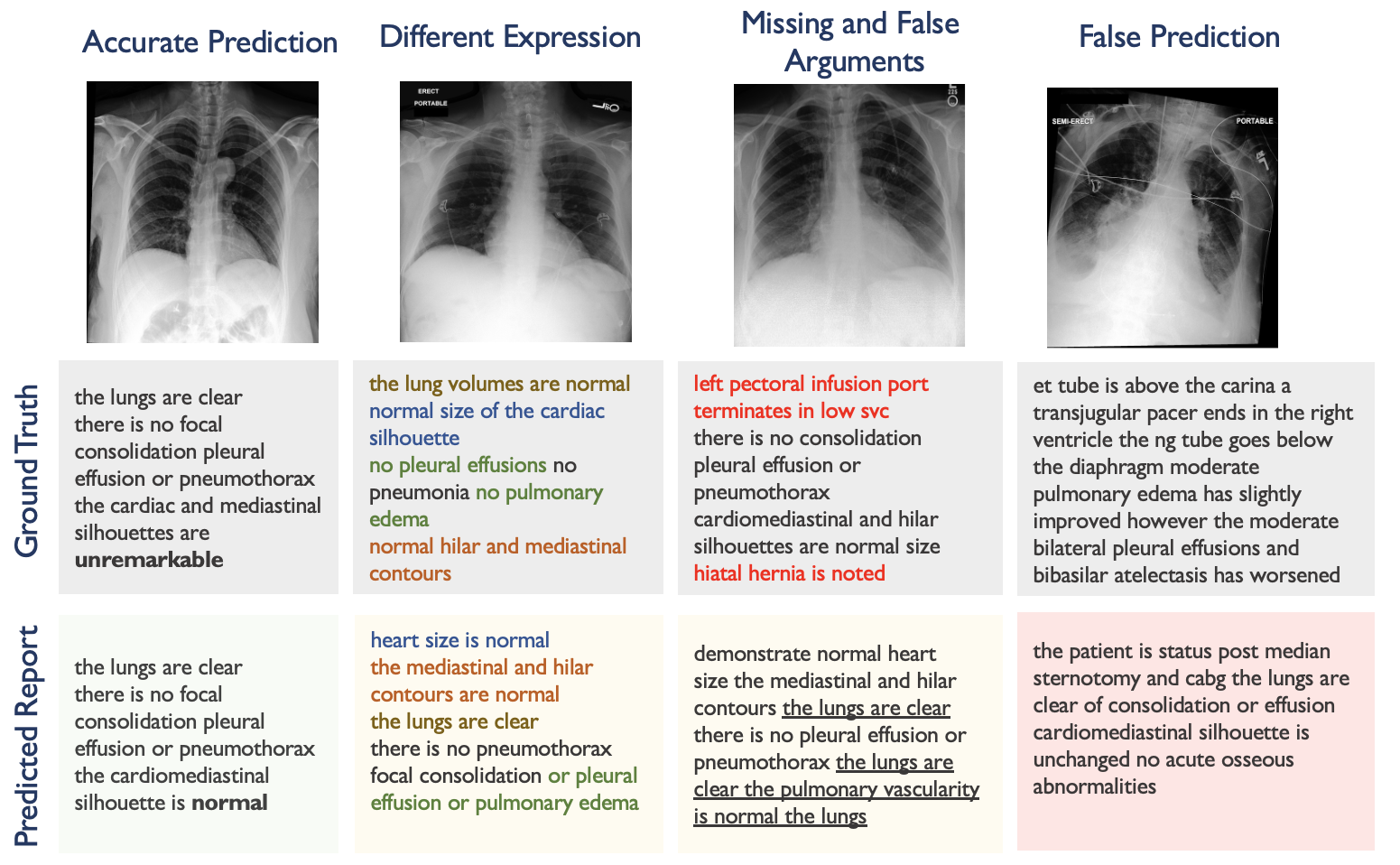}
\end{tabular}
\end{center}
\caption[results] 
{ \label{fig:results} 
Illustrations of reports from ground-truth and ethnicity-enriched models for different cases.}
\end{figure}

The example predictions from the ethnicity-enriched model are shown in Fig.~\ref{fig:results}. For a better assessment of the results, we classified the illustrated samples into four categories: accurate prediction, different expressions, missing and false arguments, and false prediction. The text that conveys the same meaning is shown in the same colour. The red text represents missing statements, whereas the underlined text represents incorrect predictions. In all cases, the order of findings aligns with the reports written by the radiologists and the generated reports are structurally correct. On the other hand, the model was not able to identify rare cases in the test set. It also missed some arguments and generated redundant and repetitive words. Even though we tried to balance the train set, the number of some reports was relatively more than others which caused a bias while generating the next word.



\section{CONCLUSION}
This paper introduced a multi-modal transformer network that is capable of using both imaging and non-imaging data (metadata) to generate radiology reports for CXR images. The ethnicity-enriched model achieved the highest scores followed by the gender- and age-enriched ones. Using two types of metadata (gender and ethnicity) simultaneously did not show any improvement over the previous one-type approaches. The proposed approach achieves 6.1\%, 8.0\%, 1.4\%, 11\% and 17.4\% improvement over the baseline, in terms of BLEU-1, BLEU-2, BLEU-3, $P_{BERT}$ and $F1Score_{BERT}$, respectively.

\section{Acknowledgements}
 
This study is fully sponsored by Turkish Ministry of National Education. 
 
 \newpage
\bibliography{report} 
\bibliographystyle{spiebib} 

\end{document}